\documentclass[10pt,twocolumn,letterpaper]{article}

\usepackage[pagenumbers]{cvpr}

\definecolor{cvprblue}{rgb}{0.21,0.49,0.74}
\usepackage[breaklinks,colorlinks,allcolors=cvprblue]{hyperref}

\usepackage{amsmath}
\usepackage{amsfonts}

\usepackage[table]{xcolor}
\definecolor{propbg}{RGB}{252,241,232}
\definecolor{openbg}{RGB}{233,242,250}
\definecolor{ourbg}{RGB}{236,248,243}
\definecolor{improvegreen}{RGB}{0,145,0}
\usepackage{multirow}

\title{LUT: Latent Utility Training for Visual Reasoning}

\author{
Jiaxuan Kang, Siyu Chen, Mingda Li, Mingjie Liu, Tianyue Wang,\\
Zhaoyang Wei, Yongheng Zhang, Yanchao Hao, Zheng Wei, Shaocheng Wu\\
Tencent PCG
}

\begin{document}
\maketitle

\begin{abstract}
Latent visual reasoning methods have shown promise in improving performance on perception-intensive tasks. Existing latent visual reasoning methods often rely on additional intermediate supervision, while a few recent studies have begun to explore learning latent representations using only final-answer supervision. However, such coarse-grained supervision does not directly specify which latent trajectories should be prioritized for distillation or how optimization signals should be allocated across latent steps. To address these challenges, we propose LUT, a latent utility training framework that uses only standard VQA-format data to guide latent distillation and policy optimization at two levels. At the trajectory level, Utility-Aware Latent Distillation SFT induces teacher latent trajectories, ranks and filters them using relative answer-likelihood gain, and distills the selected trajectories through curriculum learning. At the step level, Latent Attribution Policy Optimization leverages answer-to-latent attention to estimate step relevance and assign differentiated weights to latent policy updates. Experiments on perception-intensive visual reasoning benchmarks demonstrate strong performance among latent-only reasoning methods and competitive results against latent-text interleaved methods, without requiring additional intermediate annotations.

\end{abstract}

\section{Introduction}

Multimodal large language models \citep{zhang2023multimodal, gemini-2.5, internvl3, hong2025glm} have advanced visual understanding, yet perception-intensive visual reasoning \citep{huang2025vision, cheng2026visual, qiang2025ver, wei2026seeing} remains challenging. Such tasks require models to identify and integrate relevant visual evidence before generating answers. Latent visual reasoning offers a promising approach by performing intermediate computations in continuous hidden states.

\begin{figure}[t]
    \centering
    \includegraphics[width=\columnwidth]{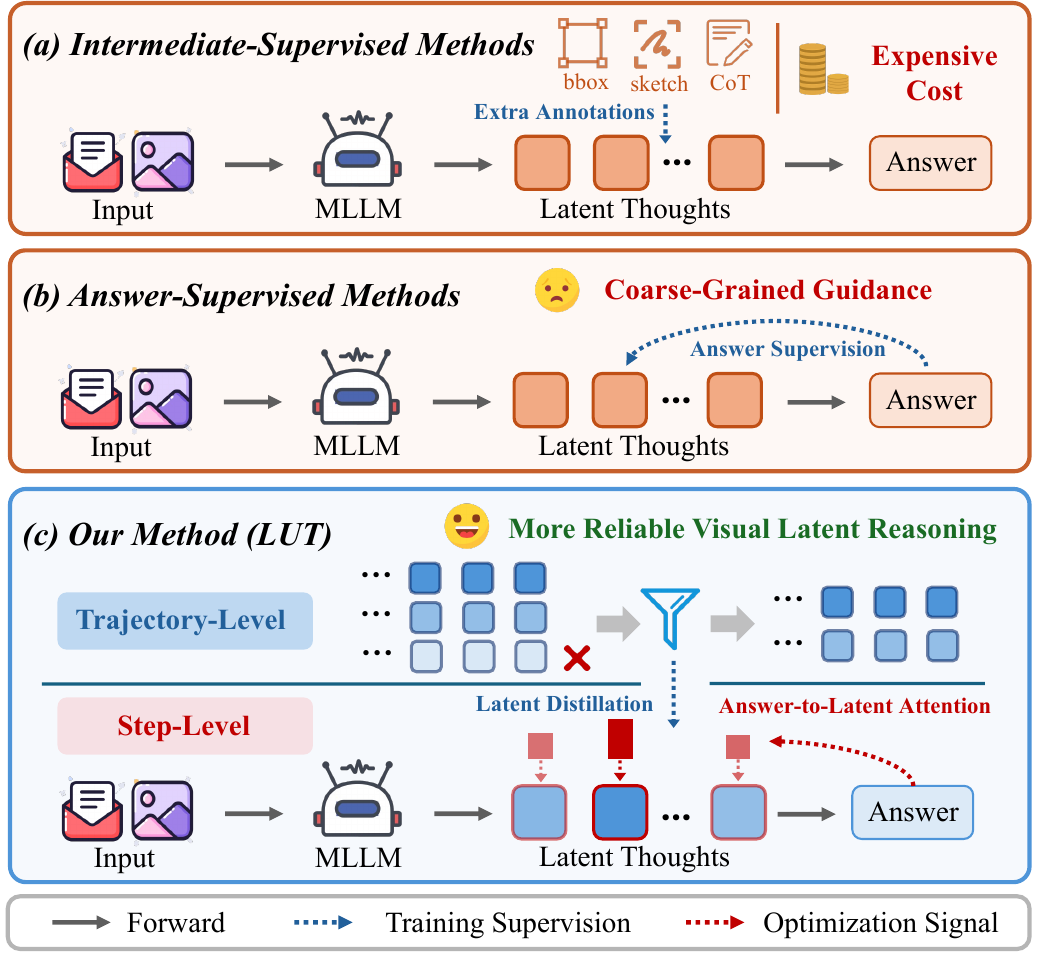}
    \caption{
    Comparison of latent visual reasoning paradigms.
    (a) Intermediate supervision requires additional annotations.
    (b) Answer supervision lacks explicit latent-level guidance.
    (c) LUT guides trajectory-level distillation and step-level optimization using answer-conditioned signals.
    }
    \label{fig:intro}
    \vspace{-8pt}
\end{figure}

Recent latent visual reasoning methods employ various forms of intermediate supervision. \citet{mirage} distill latent visual tokens from synthetic auxiliary images, while \citet{lvr} use region-level supervision to reconstruct question-relevant visual representations. \citet{monet} and \citet{hylar} further integrate latent states into interleaved visual and textual reasoning trajectories. These approaches often require additional supervision targets or structured reasoning traces, which increase data preparation costs and impose predefined forms on intermediate reasoning, as illustrated in Figure~\ref{fig:intro}(a). More recently, a few studies have begun to explore learning latent visual representations using only final-answer supervision \citep{livr}. Figure~\ref{fig:intro}(b) illustrates this emerging direction, which reduces the dependence on explicit intermediate annotations.

However, although final-answer supervision can support latent learning, its training signals remain coarse-grained and provide limited direct guidance for intermediate latent representations. At the \textbf{trajectory level}, latent trajectories induced from answer supervision may differ in their suitability for distillation, yet there is no explicit supervision to determine which trajectories should be selected or prioritized. At the \textbf{step level}, even when continuous latent states are directly optimized through reinforcement learning, a shared sequence-level advantage does not explicitly distinguish the optimization priorities of individual latent steps. Therefore, how to selectively utilize final-answer supervision for effective latent learning and optimization remains an important challenge.

To address this challenge, we propose \textbf{LUT}, \textbf{Latent Utility Training for Visual Reasoning}, a framework trained using only standard VQA-format data consisting of image-question-answer triples. As illustrated in Figure~\ref{fig:intro}(c), LUT adopts \textbf{Latent Utility} as a unified training principle, using answer-conditioned signals to guide supervision selection at the trajectory level and optimization signal allocation at the step level.

At the trajectory level, we propose \textbf{Utility-Aware Latent Distillation SFT}, which follows a three-stage pipeline. First, a visual bottleneck mask guides a teacher model to induce task-relevant latent trajectories. Second, the trajectories are scored and filtered based on their relative answer-likelihood gain, providing a utility ordering for subsequent distillation. Finally, the selected trajectories are distilled into a student initialized from the base model through a low-to-high utility curriculum. This process enables more selective use of latent supervision derived from final answers. At the step level, we propose \textbf{Latent Attribution Policy Optimization (LAPO)}, using answer-to-latent attention to estimate the relative relevance of individual latent steps to answer generation. Based on these estimates, LAPO assigns differentiated optimization weights to continuous latent actions, enabling more selective latent policy optimization.

Experiments on perception-intensive visual reasoning benchmarks demonstrate the effectiveness of LUT without additional intermediate annotations. Ablation studies further support the benefits of both training stages. Our main contributions are summarized as follows:

\begin{itemize}
    \item We introduce \textbf{LUT}, a latent visual reasoning framework that adopts latent utility as a unified training principle. Using only standard VQA-format data, LUT guides latent supervision selection and optimization signal allocation at the trajectory and step levels.

    \item We propose \textbf{Utility-Aware Latent Distillation SFT}, which induces, scores, and selects teacher latent  trajectories based on relative answer-likelihood gain, and distills them into a student through curriculum learning.

    \item We propose \textbf{LAPO}, which leverages answer-to-latent attention to assign differentiated policy optimization weights across continuous latent steps, improving latent visual reasoning performance.
\end{itemize}

\begin{figure*}[t]
	\centering
	\includegraphics[width=0.98\textwidth]{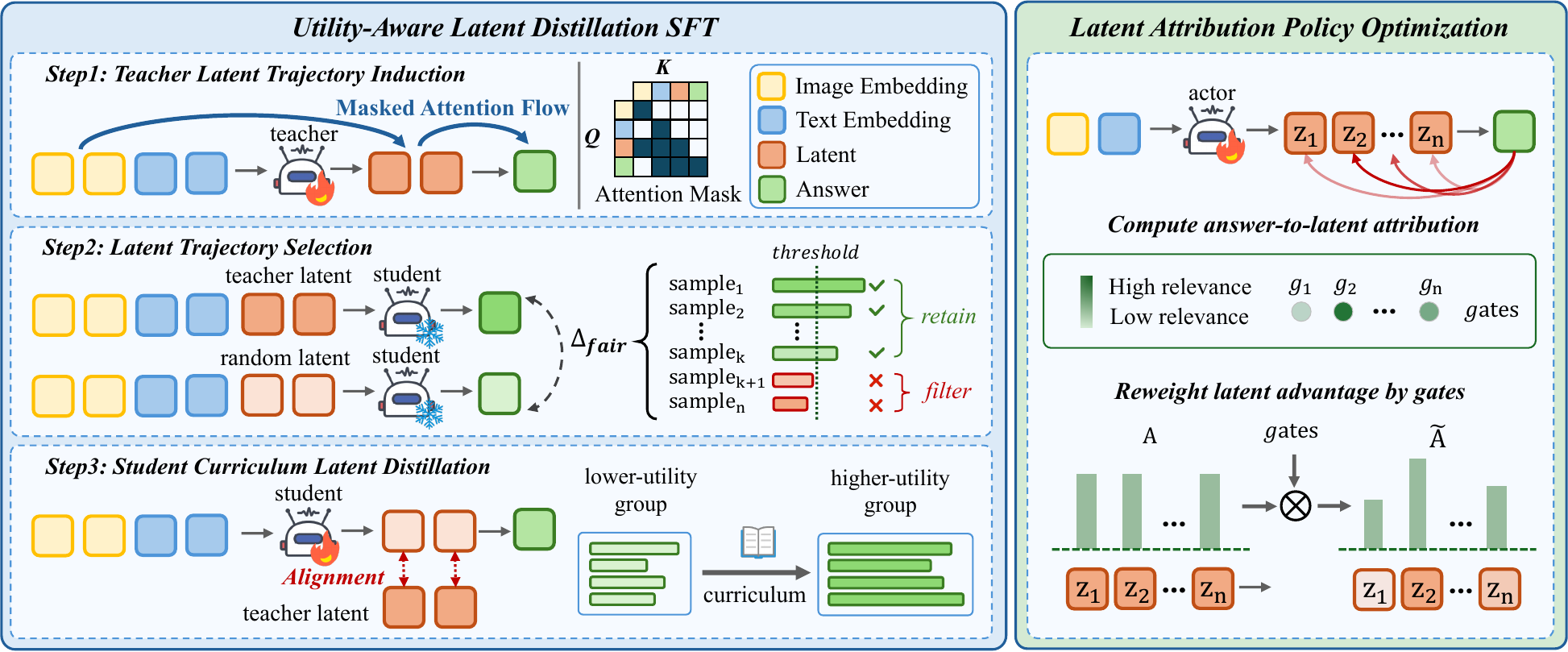}
    \caption{
    Overview of LUT.
    The left part illustrates Utility-Aware Latent Distillation SFT, which induces teacher latent trajectories, selects them using relative answer-likelihood gain, and distills the selected trajectories into a student through curriculum learning.
    The right part illustrates LAPO, which uses answer-to-latent attention to compute step-specific gates for reweighting latent policy updates.
    }
	\label{fig:method}
    \vspace{-10pt}
\end{figure*}

\section{Related Work}

\subsection{MLLMs for Visual Reasoning}

\citet{visual-cot}, \citet{xu2025visual}, \citet{wei2025advancing}, \citet{wei2025ad}, \citet{zhang2026chatbotdigitalcolleagueparadigm}, and \citet{jiang2026vlm} improve multimodal reasoning through textual Chain-of-Thought, structured reasoning traces, and reinforcement learning. However, reasoning primarily in language space may limit the representation of fine-grained visual details, such as object localization, local comparisons, and spatial relationships. \citet{wang2025pixel}, \citet{deepeyes}, \citet{thyme}, \citet{chern2025thinking}, \citet{xu2025visual}, \citet{zhang2025vitcot}, and \citet{wei2025guidinginnereyeframework} introduce explicit visual intermediates through image manipulation, auxiliary images, or interleaved reasoning. These approaches may require additional data preparation, external tools, or predefined operations. In contrast, \citet{mcout}, \citet{lavit}, and \citet{ivt-lr} explore latent reasoning as an alternative that performs intermediate computations in continuous hidden space without fully expressing them in text.

\subsection{Latent Space Reasoning}

Latent space reasoning performs intermediate computations in continuous hidden states rather than discrete text. In language reasoning, \citet{coconut} and \citet{codi} explore continuous latent thoughts for implicit reasoning. In multimodal reasoning, \citet{lvr} use bounding boxes to supervise the reconstruction of question-relevant visual representations, while \citet{mirage} and \citet{skila} learn latent visual representations from auxiliary images and sketches, respectively. These methods guide latent learning through additional intermediate supervision.

Recent studies further explore diverse latent visual reasoning architectures. \citet{monet}, \citet{valr}, \citet{hylar}, and \citet{scolar} integrate continuous visual latents into textual reasoning trajectories. \citet{univlr} combine textual reasoning and auxiliary visual evidence during training while performing latent-only reasoning at inference. Other approaches investigate alternative latent learning mechanisms \citep{gap,pearl,prolavit,amvl}. 
More recently, \citet{livr} learn a shared set of additional embeddings from final-answer supervision through visual bottleneck training, but do not further investigate how to assess the reliability of weakly supervised latent trajectories or differentially optimize individual latent steps. In contrast, LUT estimates relative trajectory utility from answer likelihood to guide trajectory selection and curriculum distillation.

Beyond latent representation learning, \citet{monet} introduce VLPO to optimize continuous latent actions, while \citet{lto} explore latent reward modeling for language reasoning. \citet{realvr} further investigate visual evidence grounding and latent credit assignment. In contrast, LUT uses answer-to-latent attention to assign differentiated policy optimization weights across continuous latent steps without additional intermediate annotations.

\section{Method}

\subsection{Overall Framework}
We propose LUT, a latent utility training framework that learns from standard VQA-format data without additional intermediate annotations. We regard latent utility as an answer-oriented training principle that guides the selection and allocation of latent training signals. Specifically, LUT uses answer-conditioned proxy signals to prioritize latent supervision at the trajectory level and differentiate policy updates at the step level.

\paragraph{Inference.}

Given an image \(I\) and a question \(Q\), LUT autoregressively generates a fixed-length latent trajectory before producing the answer:
\[
\text{\texttt{<latent>}},\ z_1,\ldots,z_N,\ \text{\texttt{</latent>}}
\]
where \(z_i\in\mathbb{R}^d\) denotes the continuous hidden state at the \(i\)-th latent step, and \(N\) is the predefined latent budget. The model then generates the final answer conditioned on the input and the entire latent trajectory.

\paragraph{Training.}

As illustrated in Figure~\ref{fig:method}, LUT consists of two training stages. First, Utility-Aware Latent Distillation SFT induces teacher latent trajectories, selects them using relative utility estimates, and distills the selected trajectories into a student through curriculum learning, as detailed in Section~\ref{sec:sft}. Second, Latent Attribution Policy Optimization (LAPO) uses answer-to-latent associations to assign differentiated optimization weights across latent steps, as detailed in Section~\ref{sec:lapo}.

\subsection{Utility-Aware Latent Distillation SFT}
\label{sec:sft}

Without explicit intermediate annotations, teacher-generated latent trajectories may differ in their suitability for student distillation. To selectively utilize these trajectories, we introduce a three-stage pipeline consisting of teacher latent trajectory induction, utility-based trajectory selection, and student curriculum latent distillation.

\paragraph{Teacher Latent Trajectory Induction.}

We first train a teacher model to induce task-relevant latent trajectories from answer supervision. Following the visual bottleneck principle \citep{livr}, we apply an attention mask that prevents answer tokens from directly attending to image tokens. Question tokens are also prevented from attending to image tokens to avoid indirect visual information leakage. This encourages the teacher to encode answer-relevant visual information into latent states before generating the answer.

The teacher is optimized with answer cross-entropy and a decorrelation regularizer on the final-layer latent hidden states. Given \(Z=[z_1,\ldots,z_N]\in\mathbb{R}^{N\times d}\), the regularizer penalizes pairwise similarity between normalized latent states, encouraging diversity across latent steps. We define

\[
\hat z_i=\frac{z_i}{\|z_i\|_2},
\qquad
\mathcal{L}_{\mathrm{dec}}
=
\frac{1}{N(N-1)}
\sum_{i\neq j}
(\hat z_i^\top \hat z_j)^2 
\]
The teacher is trained with
\[
\mathcal{L}_{\mathrm{teacher}}
=
\mathcal{L}_{\mathrm{CE}}^{\mathrm{ans}}
+
\lambda_{\mathrm{dec}}\mathcal{L}_{\mathrm{dec}}
\]
This stage induces task-relevant latent trajectories from ordinary VQA supervision and provides candidate supervision for student training.

\paragraph{Latent Trajectory Selection.}

Since teacher latent trajectories are induced from answer supervision without explicit intermediate targets, their relevance to answer prediction may vary. We therefore introduce a relative answer-likelihood score to estimate their task-conditioned value and guide trajectory selection and ordering for subsequent distillation.

We use a frozen copy of the pretrained base MLLM, which also serves as the student initialization, to evaluate each teacher trajectory against a random control. For each sample, we inject either the teacher latent states \(Z^T\) or random states \(Z^R\) at the latent positions. The random states are independently sampled and rescaled to match the per-step norms of \(Z^T\), reducing perturbations caused by magnitude differences. Under teacher forcing, the frozen model computes the mean answer token log-probability:

\[
S(Z)=\frac{1}{|Y|}
\sum_{t=1}^{|Y|}
\log p_\phi(y_t\mid I,Q,Z,y_{<t})
\]
where \(\phi\) denotes the frozen original student model. The trajectory-level utility score is
\[
\Delta_{\mathrm{fair}}=S(Z^T)-S(Z^R).
\]

A larger \(\Delta_{\mathrm{fair}}\) indicates that the teacher trajectory provides stronger support for the ground-truth answer than the norm-matched random control under the frozen model. We retain trajectories whose scores exceed a predefined threshold and rank them accordingly. This score serves as a proxy for guiding supervision selection rather than a direct measure of causal utility.

\paragraph{Student Curriculum Latent Distillation.}

We initialize the student from the pretrained base MLLM rather than the teacher checkpoint. Since the teacher is optimized under a restricted attention mask, directly inheriting its parameters may introduce a mismatch with the standard causal attention used during student training. Independent initialization allows the student to learn answer prediction and latent alignment under standard attention while transferring the teacher's latent representations through distillation.

The student is trained with answer cross-entropy and all-layer cosine alignment. Let $h^{S}_{\ell,i}$ and $h^{T}_{\ell,i}$ denote the student and teacher hidden states at layer $\ell$ and latent step $i$. We define
\[
\mathcal{L}_{\mathrm{align}}
=
\frac{1}{LN}
\sum_{\ell=1}^{L}
\sum_{i=1}^{N}
\left[
1-\cos\left(h^{S}_{\ell,i},h^{T}_{\ell,i}\right)
\right],
\]
and optimize
\[
\mathcal{L}_{\mathrm{student}}
=
\mathcal{L}_{\mathrm{CE}}^{\mathrm{ans}}
+
\lambda_{\mathrm{align}}\mathcal{L}_{\mathrm{align}}.
\]

To organize the selected supervision, we adopt a curriculum based on the relative utility scores. Within each data subset, we divide the retained trajectories at the median \(\Delta_{\mathrm{fair}}\) into lower-utility and higher-utility groups, preserving similar category proportions across stages. The student is trained on the lower-utility group before the higher-utility group, with samples randomly shuffled within each group. This ordering progressively exposes the student to teacher trajectories with different levels of answer-likelihood gain, providing a structured way to utilize the selected latent supervision.

\subsection{Latent Attribution Policy Optimization}
\label{sec:lapo}

Utility-Aware Latent Distillation SFT establishes the model's latent reasoning ability through teacher distillation and answer supervision. To further optimize continuous latent actions, we introduce Latent Attribution Policy Optimization (LAPO). LAPO addresses the limited step-level guidance of sequence-level feedback by assigning differentiated optimization weights based on answer-to-latent associations.

\paragraph{Preliminary.}

Standard GRPO \citep{grpo} constructs its policy objective using discrete text token probabilities, without explicitly modeling the likelihood of continuous latent actions. \citet{monet} address this limitation through Visual-latent Policy Optimization (VLPO), which models continuous latent actions with a Gaussian policy to enable direct policy optimization.

For a latent action \(z_t\) sampled during rollout, VLPO models its likelihood under the current policy using an isotropic Gaussian with mean \(\mu_\theta(s_t)\) and fixed variance \(\sigma^2\mathbf{I}\):
\[
\log \pi_\theta(z_t\mid s_t)
=
-\frac{\|z_t-\mu_\theta(s_t)\|_2^2}{2\sigma^2}
+C,
\]
where \(C\) is independent of \(\theta\). The corresponding latent importance ratio is
\[
r_t^{\mathrm{lat}}
=
\exp\left(
\log \pi_\theta(z_t\mid s_t)
-
\log \pi_{\theta_{\mathrm{old}}}(z_t\mid s_t)
\right).
\]
LAPO adopts this formulation and further differentiates the optimization weights assigned to individual latent steps.

\begin{table*}[t]

\caption{
Performance comparison on perception-intensive visual reasoning benchmarks.
Extra Annotation denotes externally provided intermediate supervision beyond standard VQA-format data.
$^{*}$ denotes LVR results evaluated using the officially released checkpoint trained on the full 438K Visual-CoT data. $^{\dagger}$ denotes our reproduction using the same 60K Visual-CoT samples and matched training settings as LUT-7B-SFT.
Avg denotes the mean of the overall scores across the four benchmarks.
}
\label{tab:main}
\centering
\small
\setlength{\tabcolsep}{1.0pt}
\renewcommand{\arraystretch}{0.96}

\resizebox{\textwidth}{!}{
\begin{tabular}{lcccccccccccccc}
\toprule
\multirow{2}{*}{\textbf{Model}}
& \multirow{2}{*}{\shortstack{\textbf{Extra}\\\textbf{Annotation}}}
& \multicolumn{3}{c}{\textbf{VStar}}
& \multicolumn{3}{c}{\textbf{HRBench4K}}
& \multicolumn{3}{c}{\textbf{HRBench8K}}
& \multicolumn{3}{c}{\textbf{MME-RealWorld-Lite}}
& \multirow{2}{*}{\textbf{Avg}} \\
\cmidrule(lr){3-5}
\cmidrule(lr){6-8}
\cmidrule(lr){9-11}
\cmidrule(lr){12-14}
& 
& Overall & Attribute & Spatial
& Overall & FSP & FCP
& Overall & FSP & FCP
& Overall & Reasoning & Perception
& \\
\midrule

\rowcolor{gray!10}
\multicolumn{15}{c}{\textit{\textbf{Proprietary Model}}} \\

GPT-4o
& --
& 67.5 & 72.2 & 60.5
& 59.0 & 70.0 & 48.0
& 55.5 & 62.0 & 49.0
& 52.0 & 48.3 & 54.4
& 58.50 \\

\midrule

\rowcolor{gray!10}
\multicolumn{15}{c}{\textit{\textbf{Open-Source Model}}} \\

Qwen2.5-VL-7B
& No
& 76.44 & 77.39 & 75.00
& 67.00 & 83.50 & 50.50
& 64.38 & 76.50 & 52.25
& 43.67 & 36.00 & 48.59
& 62.87 \\

\quad + vanilla SFT
& No
& 81.15 & 84.35 & 76.32
& 70.00 & 83.25 & 56.75
& 64.50 & 75.75 & 53.25
& 51.22 & 46.53 & 54.23
& 66.72  \\

\quad + vanilla SFT+GRPO
& No
& 83.77 & \textbf{86.09} & 80.26
& 72.25 & 86.00 & 58.75
& 66.88 & 77.50 & 55.25
& 52.63 & 46.13 & 57.49
& 68.88  \\

LLaVA-OneVision-7B 
& --
& 71.25 & 73.48 & 67.89
& 62.50 & 74.25 & 50.75
& 58.25 & 67.50 & 49.00
& --    & --    & --
& -- \\

DeepEyes
& BBox
& 83.25 & 84.35 & 81.58
& 71.25 & 83.75 & 58.75
& 65.13 & 77.00 & 53.25
& 54.28 & \textbf{50.53} & 56.63
& 68.48 \\

\midrule

\rowcolor{gray!15}
\multicolumn{15}{c}{\textit{\textbf{Latent Reasoning Model}}} \\

LVR$^{*}$
& BBox
& 82.2  & 81.74 & \textbf{82.89}
& 70.88 & 83.75 & 58.00
& 66.75 & 77.75 & 55.75
& 50.55 & 44.67 & 54.32
& 67.60 \\

LVR$^{\dagger}$ (reproduced)
& BBox
& 79.58 & -- & --
& 70.62 & -- & --
& 65.88 & -- & --
& 51.28 & -- & --
& 66.84 \\

SkiLa-V
& Sketch
& 79.6  & -- & --
& 68.5  & -- & --
& 65.2  & -- & --
& 50.2  & 47.7 & 51.7
& 65.88 \\

UniVLR
& Aux. Images
& 82.70 & 83.50 & 81.60
& \textbf{73.30} & 86.00 & \textbf{60.50}
& \textbf{68.80} & 78.80 & \textbf{58.80}
& 50.70 & 44.70 & 54.50
& 68.88 \\

Laser
& ScanPath
& -- & -- & --
& 72.5 & -- & --
& -- & -- & --
& -- & -- & --
& -- \\

LIVR-7B
& No
& 80.1 & -- & --
& -- & -- & --
& -- & -- & --
& -- & -- & --
& -- \\

\midrule

\rowcolor{cyan!15}
\multicolumn{15}{c}{\textit{\textbf{Our Model}}} \\

\textbf{LUT-7B-SFT}
& \textbf{No}
& 80.63 & 82.61 & 77.63
& 71.75 & 85.00 & 58.50
& 67.63 & 77.00 & 58.25
& 53.78 & 48.27 & 57.31
& 68.45 \\

\textbf{LUT-7B}
& \textbf{No}
& \textbf{84.29} & \textbf{86.09} & 81.58
& 72.62 & \textbf{86.50} & 58.75
& 68.12 & \textbf{80.00} & 56.25
& \textbf{54.82} & 47.33 & \textbf{59.62}
& \textbf{69.96} \\

$\Delta$ (vs Qwen2.5-VL-7B)
&   
& \textcolor{green!50!black}{$\uparrow$7.85}
& \textcolor{green!50!black}{$\uparrow$8.70}
& \textcolor{green!50!black}{$\uparrow$6.58}
& \textcolor{green!50!black}{$\uparrow$5.62}
& \textcolor{green!50!black}{$\uparrow$3.00}
& \textcolor{green!50!black}{$\uparrow$8.25}
& \textcolor{green!50!black}{$\uparrow$3.74}
& \textcolor{green!50!black}{$\uparrow$3.50}
& \textcolor{green!50!black}{$\uparrow$4.00}
& \textcolor{green!50!black}{$\uparrow$11.15}
& \textcolor{green!50!black}{$\uparrow$11.33}
& \textcolor{green!50!black}{$\uparrow$11.03}
& \textcolor{green!50!black}{$\uparrow$7.09} \\

\bottomrule
\end{tabular}
}

\vspace{-4pt}
\end{table*}

\begin{figure*}[t]
    \centering
    \includegraphics[width=0.98\textwidth]{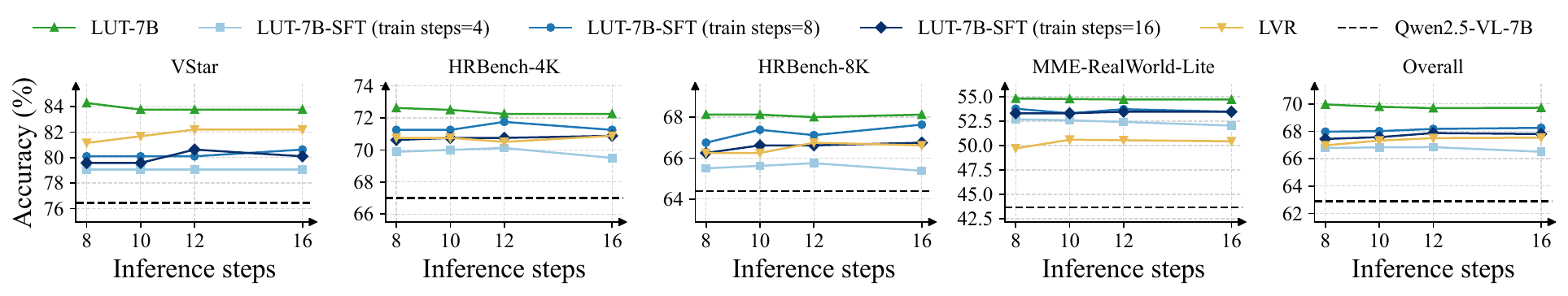}
    \caption{
    Effect of latent budgets during training and inference.
    We compare LUT-7B-SFT models trained with different latent budgets and evaluate them using inference budgets in $\{8,10,12,16\}$.
    LUT-7B is trained with the default 8-step budget.
    }
    \label{fig:latent_steps}
    \vspace{-4pt}
\end{figure*}

\begin{table*}[t]
\caption{
Comparison with latent-text interleaved reasoning methods on perception-intensive visual reasoning benchmarks.
Extra Annotation denotes externally provided intermediate supervision beyond standard VQA-format data.
Avg Tokens denotes the average number of tokens generated during inference on each benchmark.
Avg denotes the mean of the overall scores across the four benchmarks.
}
\label{tab:interleaved_comparison}
\centering
\small
\setlength{\tabcolsep}{2.6pt}
\renewcommand{\arraystretch}{0.96}
\resizebox{\textwidth}{!}{
\begin{tabular}{lcccccccccc}
\toprule
\multirow{2}{*}{\textbf{Model}}
& \multirow{2}{*}{\shortstack{\textbf{Extra}\\\textbf{Annotation}}}
& \multicolumn{2}{c}{\textbf{VStar}}
& \multicolumn{2}{c}{\textbf{HRBench4K}}
& \multicolumn{2}{c}{\textbf{HRBench8K}}
& \multicolumn{2}{c}{\textbf{MME-RealWorld-Lite}}
& \multirow{2}{*}{\textbf{Avg}} \\
\cmidrule(lr){3-4}
\cmidrule(lr){5-6}
\cmidrule(lr){7-8}
\cmidrule(lr){9-10}
&
& Overall & Avg Tokens $\downarrow$
& Overall & Avg Tokens $\downarrow$
& Overall & Avg Tokens $\downarrow$
& Overall & Avg Tokens $\downarrow$
& \\
\midrule

\rowcolor{gray!12}
\multicolumn{11}{c}{\textit{\textbf{Latent-Text Interleaved Reasoning Model}}} \\

Monet
& Aux. + CoT
& 83.25 & 85.27
& 71.00 & 85.20
& 68.00 & 76.55
& 55.50 & 134.08
& 69.44 \\

SkiLa
& Sketch + CoT
& \textbf{84.30} & 91.73
& 72.00 & 137.49
& 66.50 & 145.64
& 54.10 & 79.82
& 69.23 \\

SCOLAR-7B
& Aux. + CoT
& 83.77 & --
& \textbf{75.50} & --
& 67.63 & --
& \textbf{59.87} & --
& \textbf{71.69} \\

\midrule

\rowcolor{cyan!15}
\multicolumn{11}{c}{\textit{\textbf{Our Model (Latent Reasoning)}}} \\

\textbf{LUT-7B-SFT}
& \textbf{No}
& 80.63 & 23.42
& 71.75 & 22.50
& 67.63 & 26.49
& 53.78 & \textbf{14.00}
& 68.45 \\

\textbf{LUT-7B}
& \textbf{No}
& {84.29} & \textbf{15.14}
& 72.62 & \textbf{18.33}
& \textbf{68.12} & \textbf{18.23}
& 54.82 & \textbf{14.00}
& 69.96 \\

\bottomrule
\end{tabular}
}

\vspace{-4pt}
\end{table*}

\begin{table}[t]
\caption{
Performance comparison on broader multimodal reasoning and generalization benchmarks.
$^{*}$ denotes results using open-source official weights.
}
\label{tab:broader_reasoning}
\centering
\small
\setlength{\tabcolsep}{4pt}
\renewcommand{\arraystretch}{0.95}
\resizebox{\columnwidth}{!}{
\begin{tabular}{lccc}
\toprule
\textbf{Model}
& \textbf{MMStar}
& \textbf{BLINK}
& \textbf{VisualPuzzles} \\
\midrule

Qwen2.5-VL-7B
& 59.70
& 53.60
& 32.71 \\

LLaVA-OneVision-7B
& 59.13
& 49.34
& -- \\

DeepEyes
& 58.73
& 51.08
& 32.96 \\

LVR$^{*}$
& 57.93
& 53.60
& 29.20 \\

Laser
& 60.27
& \textbf{56.92}
& -- \\

\midrule
\rowcolor{cyan!15}

\textbf{LUT-7B (Ours)}
& \textbf{62.13}
& 55.08
& \textbf{34.67} \\

$\Delta$ (vs Qwen2.5-VL-7B)
& \textcolor{green!50!black}{$\uparrow$2.43}
& \textcolor{green!50!black}{$\uparrow$1.48}
& \textcolor{green!50!black}{$\uparrow$1.96} \\

\bottomrule
\end{tabular}
}

\vspace{-2pt}
\end{table}

\paragraph{Attribution-Guided Latent Policy Reweighting.}

Different latent steps may capture different visual cues and exhibit varying degrees of association with answer generation, as further examined in Section~\ref{sec:interpretation}. However, a shared sequence-level advantage does not explicitly distinguish their relative optimization priorities. We therefore use answer-to-latent attention as a step-level relevance proxy to guide differentiated latent policy updates.

We estimate step-level relevance using answer-to-latent attention. Let \(\mathcal{T}_{\mathrm{ans}}\) denote the positions of answer-content tokens. Over a set of final Transformer layers \(\mathcal{L}\), we average the attention probabilities from answer tokens to the \(i\)-th latent step:
\[
r_i
=
\frac{1}{|\mathcal{L}|}
\sum_{\ell\in\mathcal{L}}
\frac{1}{H|\mathcal{T}_{\mathrm{ans}}|}
\sum_{h=1}^{H}
\sum_{t\in\mathcal{T}_{\mathrm{ans}}}
\mathrm{Attn}_{\ell,h}(t,i)
\]
where \(H\) denotes the number of attention heads. We then map the relative attention score into a bounded gate using the within-sample mean \(\bar r\):
\[
g_i
=
1+\alpha\cdot
\max\left(
0,\tanh\frac{r_i-\bar r}{\tau}
\right),
\qquad
\tilde A_i=g_i A
\]
where \(A\) denotes the sequence-level advantage. The reweighted advantage \(\tilde A_i\) is applied to the corresponding latent policy update.

When \(A>0\), higher-attention latent steps receive larger positive advantage weights. When \(A<0\), these steps receive larger negative advantage weights. Thus, LAPO differentiates the optimization strength across latent steps using their answer-to-latent associations rather than assigning identical advantage weights to all steps.

\paragraph{Reward Design.}

We derive outcome rewards from answer accuracy and format compliance. The accuracy reward equals 1 if the final answer is correct and 0 otherwise. The format reward equals 1 if the response uses the required special tokens to delimit the latent trajectory and answer span, and contains the predefined number of latent steps, and 0 otherwise. These signals are used to compute the outcome reward from which the sequence-level advantage is derived.

\section{Experiments}

\subsection{Experimental Setup}

\paragraph{Training Setup.}
LUT is built on Qwen2.5-VL-7B. We retain trajectories with $\Delta_{\mathrm{fair}}>-0.5$, filtering out 6.5\% of the lowest-scoring trajectories. LAPO uses $\alpha=1.5$ and $\tau=0.5$, yielding gates in $[1.0,2.5]$. Unless otherwise specified, training uses 8 latent steps. Following \citet{monet}, we evaluate inference budgets in $\{8,10,12,16\}$. The 8-step setting achieves the best LUT-7B results across all four benchmarks and is used for the main comparison. Figure~\ref{fig:latent_steps} reports all evaluated budgets. Additional implementation details are provided in the supplementary material.

\paragraph{Training Data.}
LUT uses only VQA-format data. For SFT, we randomly sample 60K examples from Visual-CoT \citep{visual-cot} and retain only image-question-answer triples. For RL, we sample 24K examples, including 15K from ZwZ-RL-VQA \citep{zwz} with the original images and 9K from ViRL \citep{virl}. Thus, no additional intermediate annotations are used during training.

\paragraph{Evaluation Benchmarks.}
We evaluate LUT on four perception-intensive benchmarks: VStar \citep{vstar}, HRBench4K, HRBench8K \citep{hrbench}, and MME-RealWorld-Lite \citep{mme}. We further evaluate generalization on MMStar \citep{mmstar}, BLINK \citep{blink}, and VisualPuzzles \citep{visualpuzzles}. Together, these benchmarks cover fine-grained visual search, high-resolution and real-world perception, general multimodal reasoning, and out-of-distribution abstract reasoning.

\paragraph{Baselines.}
We compare LUT with general MLLMs, including GPT-4o \citep{gpt-4o}, Qwen2.5-VL-7B \citep{qwen2.5-vl}, LLaVA-OneVision-7B \citep{llava-onevision}, and DeepEyes \citep{deepeyes}. Latent-only reasoning baselines include LVR, SkiLa-V, Laser \citep{laser}, and UniVLR. Latent-text interleaved reasoning baselines include Monet, SkiLa, and SCOLAR. Results for published methods are taken from public reports unless otherwise specified. We train vanilla SFT and SFT+GRPO using the same stage-wise data as LUT. We also reproduce LVR using the same 60K Visual-CoT samples and matched SFT settings as LUT-7B-SFT, while retaining its bounding-box supervision. We also include the published VStar result of LIVR \citep{livr} for reference. Unlike LUT's autoregressively generated latent states, LIVR uses a fixed set of learnable latent input embeddings shared across samples, and its reported training and evaluation settings are not directly matched to ours.

\begin{table*}[t]
\caption{
Ablation study of Utility-Aware Latent Distillation SFT.
We vary trajectory selection, curriculum order, and teacher or student training designs.
Avg denotes the mean score across the four benchmarks.
}
\label{tab:sft_ablation}
\centering
\scriptsize
\setlength{\tabcolsep}{2.2pt}
\renewcommand{\arraystretch}{0.90}
\resizebox{\textwidth}{!}{
\begin{tabular}{lccc@{\hspace{8pt}}|@{\hspace{8pt}}ccccc}
\toprule
\textbf{Variant}
& \textbf{\(\Delta\) Threshold}
& \textbf{Retained Ratio}
& \textbf{Curriculum Order}
& \textbf{VStar}
& \textbf{HR-4K}
& \textbf{HR-8K}
& \textbf{MME-RW-Lite}
& \textbf{Avg} \\
\midrule

\textbf{LUT-7B-SFT}
& -0.5
& 93.5\%
& low-to-high
& 80.63
& 71.75
& 67.63
& 53.78
& \textbf{68.45} \\

\midrule
\rowcolor{blue!8}
\multicolumn{9}{c}{\textit{\textbf{Trajectory Selection and Curriculum Order}}} \\

w/o filtering and curriculum
& None
& 100\%
& random
& 80.63
& 70.88
& 65.88
& 53.78
& 67.79 \\

w/o curriculum
& -0.5
& 93.5\%
& random
& 81.15
& 71.25
& 66.38
& 53.26
& 68.01 \\

Stricter filtering
& 0
& 76.1\%
& low-to-high
& 81.15
& 71.88
& 67.12
& 52.79
& 68.24 \\

Reversed curriculum
& -0.5
& 93.5\%
& high-to-low
& 75.92
& 71.38
& 65.75
& 51.90
& 66.24 \\

\midrule
\rowcolor{yellow!12}
\multicolumn{9}{c}{\textit{\textbf{Teacher and Student Training Design}}} \\

w/o teacher bottleneck mask
& -0.5
& 88.0\%
& low-to-high
& 78.53
& 70.62
& 63.62
& 51.59
& 66.09 \\

w/o teacher decorrelation
& -0.5
& 94.2\%
& low-to-high
& 80.10
& 72.00
& 65.88
& 53.05
& 67.76 \\

w/o student latent alignment
& -0.5
& 93.5\%
& low-to-high
& 75.39
& 70.88
& 64.88
& 52.68
& 65.96 \\

Initialized from teacher
& -0.5
& 93.5\%
& low-to-high
& 78.01
& 67.00
& 62.62
& 51.75
& 64.85 \\

\bottomrule
\end{tabular}
}

\vspace{-5pt}
\end{table*}

\begin{table}[t]
\caption{
Ablation study of reinforcement learning strategies.
Avg denotes the mean score across the four benchmarks.
}
\label{tab:rl_ablation}
\centering
\small
\setlength{\tabcolsep}{1.2pt}
\renewcommand{\arraystretch}{1.0}
\resizebox{\columnwidth}{!}{
\begin{tabular}{lccccc}
\toprule
\textbf{Method}
& \textbf{VStar}
& \textbf{HR-4K}
& \textbf{HR-8K}
& \textbf{MME-RW-Lite}
& \textbf{Avg} \\

\midrule

LUT-7B-SFT
& 80.63
& 71.75
& 67.63
& 53.78
& 68.45  \\

\quad + GRPO
& 82.20
& 71.88
& 66.75
& 54.66
& 68.87  \\

\quad + VLPO
& 83.77
& 71.88
& 67.75
& 53.99
& 69.35   \\

\rowcolor{cyan!15}
\quad + \textbf{LAPO}
& 84.29
& 72.62
& 68.12
& 54.82
& \textbf{69.96}  \\

\quad + LAPO (reverse)
& 83.25
& 72.00
& 67.50
& 53.67
& 69.11  \\

\bottomrule
\end{tabular}
}

\vspace{-5pt}
\end{table}

\subsection{Main Results}

\paragraph{LUT achieves strong latent-only reasoning without extra intermediate annotations.}

As shown in Table~\ref{tab:main}, LUT-7B achieves the highest average performance among the compared latent-only reasoning baselines with complete results.
LUT-7B-SFT outperforms vanilla SFT, while LUT-7B further surpasses vanilla SFT+GRPO. Compared with the reproduced LVR baseline, LUT-7B-SFT achieves stronger performance without relying on bounding-box supervision.

\paragraph{LUT offers a favorable trade-off against latent-text interleaved methods.}
Table~\ref{tab:interleaved_comparison} shows that LUT outperforms Monet and SkiLa in average accuracy while generating substantially fewer tokens, and remains competitive with SCOLAR. This demonstrates a favorable trade-off between reasoning performance and output length without intermediate textual reasoning.

\paragraph{LUT generalizes beyond perception-intensive benchmarks.}
As shown in Table~\ref{tab:broader_reasoning}, LUT improves the base model across all three benchmarks and performs best on MMStar and VisualPuzzles, demonstrating strong generalization.

\paragraph{LUT benefits from a consistent latent budget.}
Figure~\ref{fig:latent_steps} compares different training and inference latent budgets. Training with 8 steps achieves the best overall performance. After LAPO, the same 8-step inference budget yields the best result across all four benchmarks and is therefore adopted for the main evaluation.

\begin{figure}[t]
    \centering
    \includegraphics[width=\columnwidth]{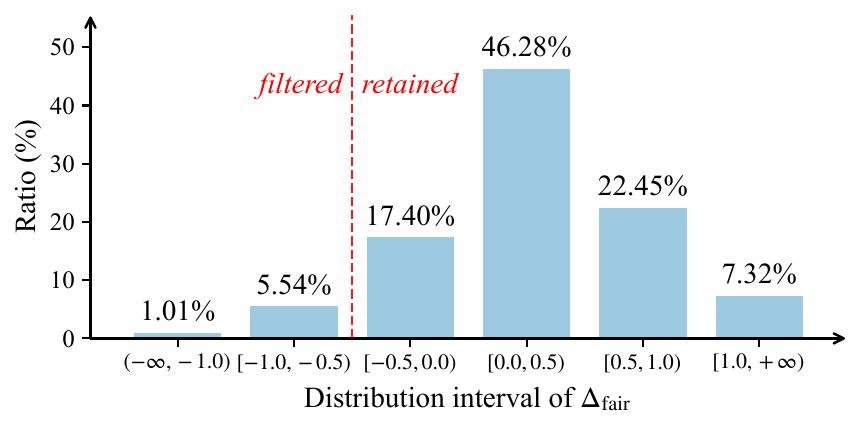}
    \caption{
    Distribution of trajectory-level utility scores $\Delta_{\mathrm{fair}}$.
    The dashed red line marks the filtering threshold.
    }
    \label{fig:delta_fair}
    \vspace{-6pt}
\end{figure}

\begin{figure}[t]
	\centering
	\includegraphics[width=\columnwidth]{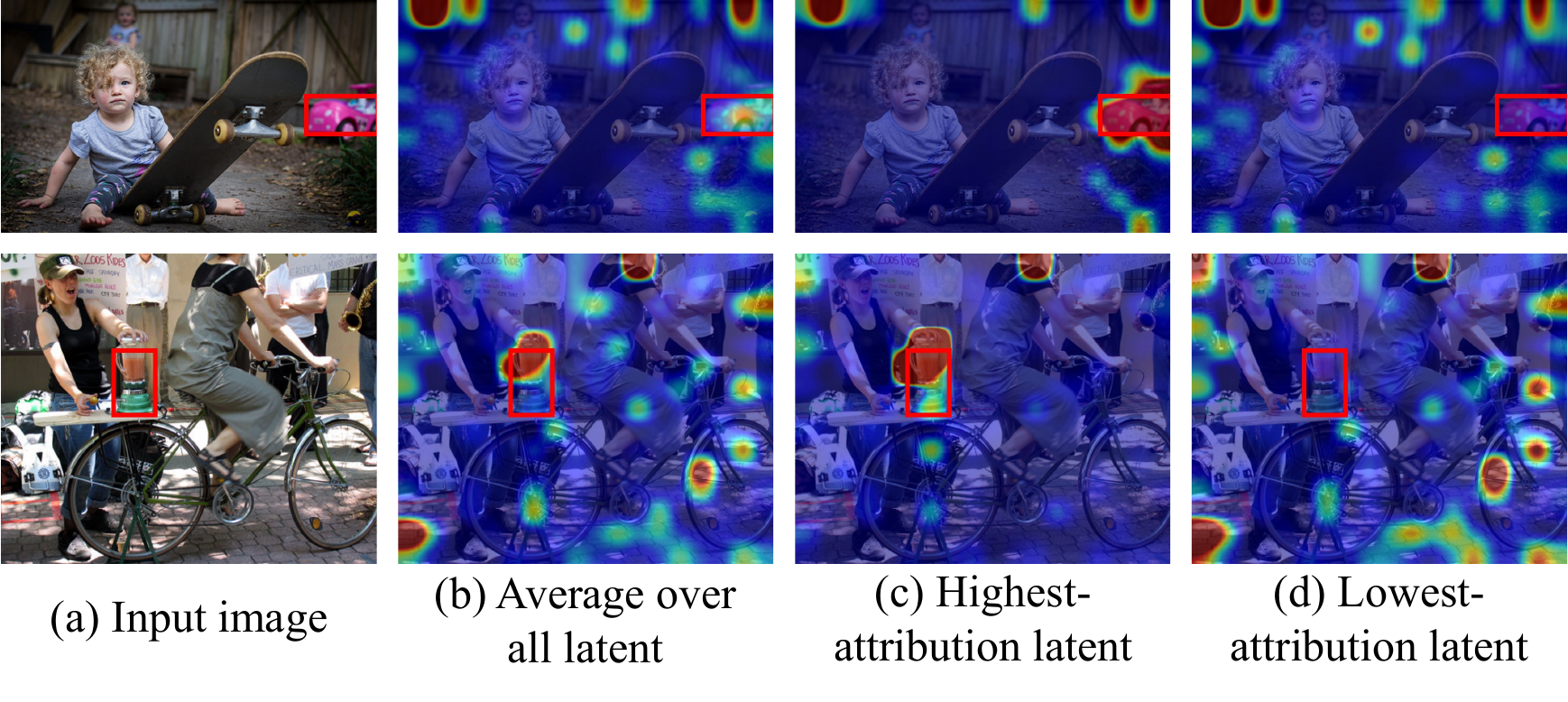}
    \caption{
    Qualitative visualization of latent-to-image attention.
    Red boxes are from the original annotations and indicate task-relevant regions.
    For each example, we show attention averaged over all latent steps, attention from the single latent step with the highest answer-to-latent attribution, and attention from the single latent step with the lowest attribution.
    }
	\label{fig:latent_interpretability}
\vspace{-6pt}
\end{figure}

\begin{figure}[t]
\centering
\includegraphics[width=0.92\columnwidth]{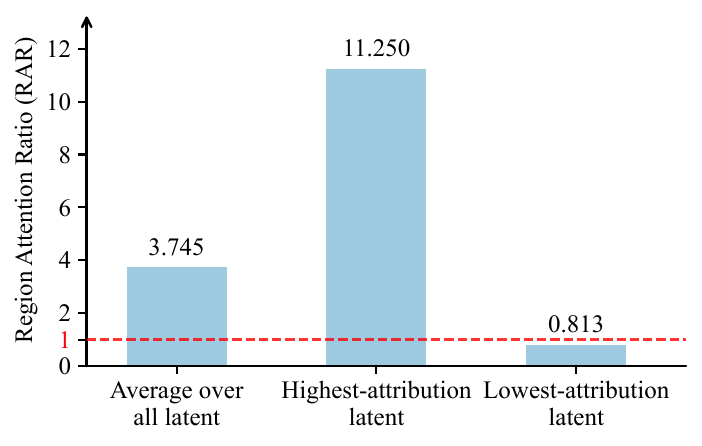}
\caption{
Average RAR of latent-to-image attention on held-out samples.
The dashed red line marks $\mathrm{RAR}=1$, where attention density inside and outside the bounding box is equal.
}
\label{fig:latent_rar}
\vspace{-8pt}
\end{figure}

\begin{table}[t]

\caption{
Effect of inference-time latent intervention.
We block answer-to-latent attention while keeping
the image and inputs unchanged.
Avg denotes the mean score across four benchmarks.
}
\label{tab:latent_intervention}

\centering
\small
\setlength{\tabcolsep}{1.2pt}
\renewcommand{\arraystretch}{1.0}
\resizebox{\columnwidth}{!}{
\begin{tabular}{lccccc}
\toprule
\textbf{Method}
& \textbf{VStar}
& \textbf{HR-4K}
& \textbf{HR-8K}
& \textbf{MME-RW-Lite}
& \textbf{Avg} \\
\midrule

\rowcolor{cyan!15}
\textbf{LUT-7B}
& 84.29
& 72.62
& 68.12
& 54.82
& \textbf{69.96}  \\

\quad w/o latent access
& 81.15 & 71.13 & 66.63 & 53.67 & 68.15 \\

\bottomrule
\end{tabular}
}
\vspace{-8pt}
\end{table}

\subsection{Ablation on SFT Stage}

\paragraph{Distribution of $\Delta_{\mathrm{fair}}$.}
Figure~\ref{fig:delta_fair} shows that most teacher trajectories obtain positive $\Delta_{\mathrm{fair}}$, indicating higher ground-truth answer likelihood than the norm-matched random control under the frozen evaluator. The threshold of $-0.5$ retains 93.5\% of trajectories, removing only the lowest-scoring samples. The effectiveness of this selection strategy is further examined through the SFT ablations.

\paragraph{Filtering and curriculum both improve trajectory distillation.}
As shown in Table~\ref{tab:sft_ablation}, removing both filtering and curriculum leads to weaker performance. Adding filtering improves the average score, suggesting that $\Delta_{\mathrm{fair}}$ provides a useful signal for selecting latent supervision. A stricter threshold performs slightly worse than the full setting, suggesting that overly aggressive filtering may reduce useful data diversity. The full low-to-high curriculum achieves the best result, while reversing the curriculum causes a clear drop. These results support the effectiveness of both trajectory selection and curriculum ordering for latent distillation.

\paragraph{Each teacher and student design contributes to performance.}
Table~\ref{tab:sft_ablation} further shows that removing the teacher bottleneck mask substantially reduces performance, indicating the importance of controlled information flow for inducing useful trajectories. Removing teacher decorrelation also hurts performance, supporting its role in latent induction. Without student latent alignment, the student relies only on answer supervision and performs worse. Initializing the student from the teacher checkpoint yields the lowest score, suggesting that independent student initialization is more effective under our distillation setting.

\subsection{Ablation on RL Stage}

\paragraph{Attribution-guided reweighting improves latent policy optimization.}
As shown in Table~\ref{tab:rl_ablation}, GRPO brings only limited gains over LUT-7B-SFT, while VLPO improves performance by directly optimizing continuous latent steps. LAPO achieves the best result, supporting the benefit of differentiated latent policy updates based on answer-to-latent associations. To examine the effect of weight allocation, we introduce LAPO (reverse), which assigns stronger updates to lower-attribution steps rather than higher-attribution ones. Its performance drops below both LAPO and VLPO, suggesting that the direction of attribution-guided reweighting matters for latent policy optimization.

\subsection{Latent Utilization and Visual Grounding}
\label{sec:interpretation}

\textbf{Latent access contributes to answer generation.}
To examine whether LUT utilizes its latent representations during inference, we block answer-to-latent attention while keeping the original image and other inputs unchanged. As shown in Table~\ref{tab:latent_intervention}, blocking latent access reduces performance across all four benchmarks, with an average drop of 1.81 percentage points. This indicates that the learned latent states contribute to answer prediction even when direct visual information remains accessible.

To further examine the visual grounding of latent representations, we conduct a post-hoc analysis on 6K held-out Visual-CoT samples. Figures~\ref{fig:latent_interpretability} and \ref{fig:latent_rar} compare latent-to-image attention averaged over all latent steps, as well as attention from the highest- and lowest-attribution steps. Red boxes denote original annotations of task-relevant regions and are used only for analysis. We quantify visual alignment using Region Attention Ratio (RAR), which compares attention density inside and outside the annotated regions. A RAR greater than 1 indicates higher attention density within the bounding box. Further details are provided in the supplementary material.

\textbf{Latent trajectories attend to task-relevant regions.}
As shown in Figure~\ref{fig:latent_interpretability}, attention averaged over all latent steps concentrates around the annotated regions. Figure~\ref{fig:latent_rar} further supports this observation, with $\mathrm{RAR}_{all}=3.745$. This suggests that LUT's latent representations exhibit attention alignment with task-relevant visual regions without bounding-box supervision during training.

\textbf{Higher-attribution steps show stronger visual alignment.}
The highest-attribution step achieves $\mathrm{RAR}_{top}=11.250$, while the lowest-attribution step has $\mathrm{RAR}_{bottom}=0.813<1$. This contrast indicates that latent steps exhibit different degrees of alignment with task-relevant visual evidence. Notably, higher answer-to-latent attribution corresponds to stronger visual alignment, supporting its use as a step-level relevance proxy in LAPO.

\section{Conclusion}

We present \textbf{LUT}, a latent visual reasoning framework that makes more selective use of coarse-grained answer supervision without additional intermediate annotations. At the trajectory level, Utility-Aware Latent Distillation SFT selects and distills teacher latent trajectories through utility-guided curriculum learning. At the step level, LAPO uses answer-to-latent attention to assign differentiated optimization weights across continuous latent steps. Experiments demonstrate strong performance on perception-intensive visual reasoning benchmarks, while ablation studies and latent analyses support the effectiveness of the proposed methods. Future work will explore more efficient training pipelines for scalable answer-supervised latent reasoning.

{
    \small
    \bibliographystyle{ieeenat_fullname}
    \bibliography{references}
}

\end{document}